\documentclass[11pt,a4paper]{article}
\usepackage[hyperref]{acl2018}
\usepackage{times}
\usepackage{latexsym}
\usepackage{url}

\usepackage{amsmath,graphicx}
\usepackage{siunitx} 
\usepackage{url}
\usepackage{booktabs}
\usepackage{microtype}

\aclfinalcopy 
\def\aclpaperid{686} 

\title{Investigating Audio, Video, and Text Fusion Methods for End-to-End Automatic Personality Prediction}
\author{Onno Kampman, Elham J. Barezi, Dario Bertero, Pascale Fung \\ 
  Center for AI Research (CAiRE)\\
  Hong Kong University of Science and Technology, Clear Water Bay, Hong Kong\\
  {\tt ejs,dbertero@connect.ust.hk, pascale@ece.ust.hk} \\}

\date{}

\begin{document}
\maketitle
\begin{abstract}
We propose a tri-modal architecture to predict Big Five personality trait scores from video clips with different channels for audio, text, and video data. For each channel, stacked Convolutional Neural Networks are employed. The channels are fused both on decision-level and by concatenating their respective fully connected layers. It is shown that a multimodal fusion approach outperforms each single modality channel, with an improvement of 9.4\% over the best individual modality (video). Full backpropagation is also shown to be better than a linear combination of modalities, meaning complex interactions between modalities can be leveraged to build better models. Furthermore, we can see the prediction relevance of each modality for each trait. The described model can be used to increase the emotional intelligence of virtual agents.
\end{abstract}
\section{Introduction}
\label{sec:intro}
Automatic prediction of personality is important for the development of emotional and empathetic virtual agents. Humans are very quick to assign personality traits to each other, as well as to virtual characters~\cite{nass1995can}. People infer personality from different cues, both behavioral and verbal, hence a model to predict personality should take into account multiple modalities including language, speech and visual cues.

Personality is typically modeled with the Big Five personality descriptors~\cite{goldberg1990alternative}. In this model an individual's personality is defined as a collection of five scores in range $[0,1]$ for personality traits Extraversion, Agreeableness, Conscientiousness, Neuroticism and Openness to Experience. These score are usually calculated by each subject filling in a specific questionnaire. 
The biggest effort to predict personality, as well as to release a benchmark open-domain personality corpus, was given by the ChaLearn 2016 shared task challenge~\cite{ponce2016chalearn}. All the best performing teams used neural network techniques. They extracted traditional audio features (zero crossing rate, energy, spectral features, MFCCs) and then fed them into the neural network~\cite{subramaniam2016bi,gurpinar2016combining,zhang2016deep}. A deep neural network should however be able to extract such features itself. \newcite{gucluturk2016deep} took a different approach by feeding the raw audio and video samples to the network. However they mostly designed the network for computer vision, and used the same architecture to audio input without any adaptation or considerations to merge modalities. The challenge was in general aimed at the computer vision community (many only used facial features), thus not many looked into what their deep network was learning regarding other modalities.

In this paper, we are interested in predicting personality from speech, language and video frames (facial features).
We first consider the different modalities separately, in order to have more understanding of how personality is expressed and which modalities contribute more to the personality definition. Then we design and analyze fusion methods to effectively combine the three modalities in order to obtain a performance improvement over each individual modality. The readers can refer to the survey by~\newcite{baltruvsaitis2018multimodal} for more information on multi-modal approaches. 

\section{Methodology} \label{sec:methodology}
\begin{figure*}[!htb]
  	\centering
  	\hspace*{-4.5cm} 
  	\includegraphics[scale=0.85]{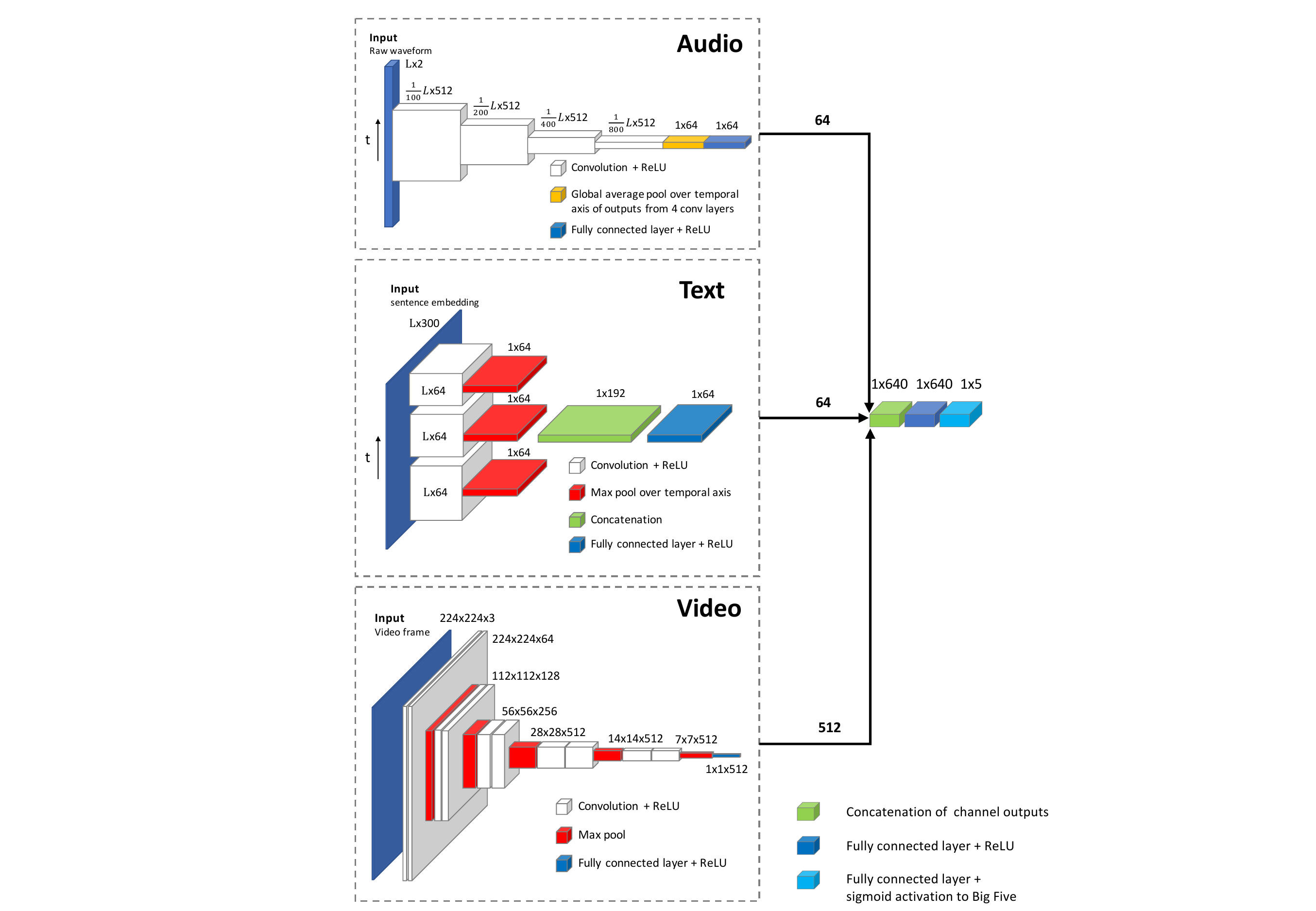}
  	\caption{\label{fig:trimodal-architecture}Diagram of the tri-modal architecture for prediction of Big Five traits from audio, text and video input. Concatenating output of the three individual modalities results in an output layer with size of $ {\scriptstyle 64+64+512=640 }$.}	
\end{figure*}

Our multimodal deep neural network architecture consists of three separate channels for audio, text, and video. The channels are fused both in decision-level fusion and inside the neural network. The three channels are trained in a multiclass way, instead of separate models for each trait~\cite{farnadi2016computational}. The full architecture is trained \textit{end-to-end}, which refers to models that are completely trained from the most raw input representation to the desired output, with all the parameters learned from data~\cite{muller2006off}. Automatic feature learning has the capacity to outperform feature engineering, as these learned features are automatically tuned to the task at hand. The full neural network architecture with the three channels is shown in Fig.~\ref{fig:trimodal-architecture}.

\subsection{Audio channel}
The audio channel looks at acoustic and prosodic (i.e. non-verbal) information of speech. It takes raw waveforms as input instead of commonly used spectrograms or traditional feature sets, as CNNs are able to autonomously extract relevant features directly from audio \cite{bertero2016emnlp}.

Input audio samples are first downscaled to a uniform sampling rate of $8\,\text{kHz}$ before fed to the model. During each training iteration (but not during evaluation), for each input audio sample we randomize the amplitude (volume) through an exponential random coefficient $\alpha = 10^{\text{U}(-1.5, 1.5)}$, where $\text{U}(-1.5, 1.5)$ is a uniform random variable, to avoid any bias related to recording volumes.

We split the input signal into two feature channels as input for the CNN: the raw waveform as-is, and the signal with squared amplitude (aimed at capturing the energy component of the signal). A stack of four convolutional layers is applied to the input to perform feature extraction from short overlapping windows, analyze variations over neighboring regions of different sizes, and combine all contributions throughout the sample.

We used global average pooling operation over the output of each layer to capture globally expressed personality characteristics over the entire audio frame and to combine the contributions of the convolutional layer outputs. After obtaining the overall vector by weighted-average of each convolutional layer output, it is fed to a fully connected layer with final sigmoid layer to perform the final regression operation to map this representation to a score in range $[0,1]$ for each of the five traits. 

\subsection{Text channel}
The transcriptions for the ChaLearn dataset, provided by the challenge organizers, were obtained by using a professional human transcription service \footnote{http://www.rev.com} to ensure maximum quality for the ground truth annotations. 
For this channel we extract {\tt word2vec} \textit{word embeddings} from transcriptions and feed those into a CNN. The embeddings have a dimensionality of $k=300$ and were pre-trained on Google News data (around 300 billion words). This enables us to take into account much more contextual information than available in just the corpus at hand.

Transcriptions per sample were split up into different sentences. We normalized the text in order to align our corpus with the embedding dictionary. We fed the computed matrix into a CNN, whose architecture is inspired by~\newcite{kim2014convolutional}. Three convolutional windows of size three, four, and five words are slid over the sentence, taking steps of one word each time. These windows are expected to extract compact $n$-grams from sentences.
After this layer, a max-pooling is taken for the outcome of each of the kernels separately to get a final sentence encoding.
The representation is then mapped through a fully connected layer with sigmoid activation, to the final Big Five personality traits.

\subsection{Video channel}

In the video channel, we first take a random frame from each of the videos, which leads to personality recognition from only appearance. We did not use Long Short Term Memory (LSTM) networks because we only need appearance information, not temporal and movement information. Although many works in the ChaLearn competition align faces manually using the famous Viola-Jones algorithm~\cite{viola2004robust} and crop them from frames~\cite{gurpinar2016combining}, here we choose not to in order to prevent excessive preprocessing. It has also been found and shown that deep architectures can automatically learn to focus on the face~\cite{gucluturk2017visualizing}.

We extract representations from the images using the VGG-face CNN model~\cite{parkhi2015deep}, with pre-trained VGG-16 weights~\cite{simonyan2014very}. Input images are fed into the model with their three channels (blue, red, and green). Several convolutional layers combined with max-pooling and padding layers follow. We use two final fully connected layers,
followed by sigmoid activation to map to the five traits. We only train these two final layers. Fine-tuning pre-trained models as such is a common way to leverage on training on large outside datasets and thus the model doesn't need to learn extracting visual features itself~\cite{esteva2017dermatologist}.

\subsection{Multimodal fusion}
Humans infer personality from different cues and this motivates us to predict personality by using multiple modalities simultaneously. We look at three different fusing methods to find how to combine modalities best.

The first method is a decision-level fusion approach, done through an ensemble (\textit{voting}) method. The three channels are copies of the fully trained models described above. We want to know the linear combination weights for each modality, for each trait (15 weights in total). The final predictions from our tri-modal model then become $\hat{p}_i = \sum_{j=1}^N w_{i,j} \hat{p}_{i,j} $, where $p_i$ represents the ensemble estimator for the score of trait $i$, $j$ represents the modality (with $N=3$ the number of modalities), and $w_{i,j}$ and $\hat{p}_{i,j}$ the weights and estimates respectively for trait $i$ for modality $j$. The weights were found by minimizing the Mean Absolute Error (MAE) on the development set. We choose to have weights per trait because the uni-modal results show that some modalities are better at predicting some traits than others. An important advantage of using this fusion method is that we can read the relevance of the modalities for each of the traits , from the weights. 

In the other fusion methods we propose, we first merge the modalities by truncating the final fully connected layers from each of the channels. We then concatenate the previous fully connected layers, to obtain shared representations of the input data. Finally we add two extra fully connected layers on top. For the second method, all layers in the separate channels are frozen, so we basically want the model to learn what combination (could still be a linear one) of channel outputs is optimal. For the third method, we again train this exact architecture, but now we also update the parameters of both the audio and text channels, thus enabling fully end-to-end backpropagation. This enables the model to learn more complex interaction between the different channels. 
The architecture is illustrated in Fig.~\ref{fig:trimodal-architecture}. The layers in the dashed boxes are frozen (non-trainable) for limited backpropation model, and trainable for full backpropagation model.

\section{Experiments} \label{sec:experiments}
\subsection{Corpus}
We used the ChaLearn First Impressions Dataset, which consists of YouTube vlogs clips of around 15 seconds~\cite{ponce2016chalearn}. The speaker in each video is annotated with Big Five personality scores. The ChaLearn dataset was divided into a training set of 6,000 clips and $20\%$ of the training set was taken as validation set during training to tune the hyperparameter, the early stopping conditions and the ensemble method training. We used pre-defined ChaLearn Validation Set of 2,000 clips as the test set. 

\subsection{Setup}
For the audio CNN, we used a window size of 200 (i.e. $25\,\text{ms}$) for the first layer, and a stride of 100 (i.e. $12.5\,\text{ms}$). In the following convolutional layers we set the window size and stride is set to 8 and 2 respectively. The number of filters was kept at 512 for each layer. In the text CNN instead we used a filter size of 128, and apply dropout ($p = 0.5$) to the last layer. In the video CNN we used again 512 filters for each layer.

We trained our model using Adam optimizer \cite{kingma2014adam}. All models were implemented with Keras \cite{chollet2015keras}.
We train all models parameters with a regression cost function by iteratively minimizing the average over five traits of the Mean Square Error (MSE) between model predictions and ground truth labels. We also use the MSE with the ground truth to evaluate the performance over the test set. 

\subsection{Results}\label{sec:results}
Our aim is to investigate the contribution of different modalities for personality detection task. 
Table~\ref{tab:tri-modal-weights} includes the optimal weights learned for the decision-level fusion approach. From this table we can read contribution of each of the modalities to the prediction of each trait.

\begin{table}[htb]
\centering
\small
\begin{tabular}{lccccc}
	\toprule
    	 			& \multicolumn{5}{c}{Big Five Personality Traits} 	\\
    \cmidrule(r){2-6}
	Model 		 	&E 		&A 		&C 		&N 		&O 		\\ 
    \midrule
	Audio			&0.44 	&0.32 	&0.27 	&0.45 	&0.54 	\\
	Text 			&-0.03 	&0.22 	&0.13 	&0.03 	&-0.06 	\\
	Video 			&0.59 	&0.46 	&0.60 	&0.52 	&0.52 	\\
    \bottomrule
\end{tabular}
\caption{\label{tab:tri-modal-weights}Optimal weights learned for combining the three modalities for each trait. E, A, C, N, and O stand for Extraversion, Agreeableness, Conscientiousness, Neuroticism and Openness, respectively.}
\end{table}

Table~\ref{tab:cnn_tri-modal_table} displays the tri-modal regression (MAE) performance, individual modalities, multimodal decision-level fusion and the two neural network fusion methods. We also report the average trait scores from the training set labels as a baseline \cite{mairesse2007using}. The neural network fusion with full backpropagation works best with an average MSE score of 0.0938, around 2.9\% improvement over the last-layer backpropagation only, and 9.4\% over the best separate modality (video). Both in separate and ensemble methods, the results we obtain are lower than just estimating the average from the training set.


\begin{table}[htb]
\centering
\small
\scalebox{0.8}{
\begin{tabular}{lcccccc}
	\toprule
    \textit{MAE}&		& \multicolumn{5}{c}{Big Five Personality Traits} 	\\
    \cmidrule(r){3-7}
	Model 		&Mean 	&E 		&A 		&C 		&N 		&O 		\\
    \midrule
	Audio		&.1059 	&.1080 	&.0953 	&.1160 	&.1077 	&.1024 	\\
	Text 		&.1132 	&.1177 	&.0977 	&.1206 	&.1167 	&.1135 	\\
	Video 		&.1035 	&.1040 	&.0960 	&.1087 	&.1064 	&.1024 	\\
    DLF			&.0967 	&.0970 	&.0893 	&.1049 	&.0979 	&.0947 	\\
	NNLB		&.0966 	&.0970 	&.0896 	&.1038 	&.0973 	&.0951 	\\
	NNFB		&.0938 	&.0958 	&.0907 	&.0922 	&.0964 	&.0938 	\\
    \cmidrule(r){2-7}
	Train labels avg	&.1165 	&.1194 	&.1009 	&.1261 	&.1209 	&.1153 	\\
    \bottomrule
\end{tabular}
}
\caption{\label{tab:cnn_tri-modal_table} MAE for each individual modality, fusion approaches and average of training set labels. DLF refers to decision-level fusion, NNLB and NNFB refer to neural network limited backprop and full back respectively.}
\end{table}

The main target in this work is to investigate the effect of audio, visual, and text modalities, and different fusion methods in personality recognition, rather than proposing the method with the best accuracy. However, we still repeat the accuracy of the reported methods in Table \ref{tab:cnn_tri-modal_table} and two winners of the ChaLearn 2016 competition DCC \cite{gucluturk2016deep} and 
evolgen \cite{subramaniam2016bi} in Table \ref{tab:chalearn_table}. It can be seen that the result for out tri-modal method with fully back-propagation is comparable to the winners.

\begin{table}[htb]
\centering
\small
\scalebox{0.8}{
\begin{tabular}{lcccccc}
	\toprule
    \textit{Mean Acc}&		& \multicolumn{5}{c}{Big Five Personality Traits} 	\\
    \cmidrule(r){3-7}
	Model 		&Mean 	&E 		&A 		&C 		&N 		&O 		\\
    \midrule
	Audio		&.8941 	&.8920 	&.9047 	&.8840 	&.8923 	&.8976 	\\
         		
	Text 		&.8868 	&.8823 	&.9023 	&.8794 	&.8833 	&.8865	\\
   			
	Video 		&.8965 	&.8960 	&.9040 	&.8913 	&.8936 	&.8976 	\\
         		
    DLF			&.9033 	&.9030 	&.9107 	&.8951 	&.9021 	&.9053 	\\
           	
	NNLB		&.9034 	&.9030 	&.9104 	&.8962 	&.9027 	&.9049 	\\
 				
	NNFB		&.9062 	&.9042 	&.9093 	&.9078 	&.9036 	&.9062 	\\

	DCC         &.9121 &.9104 &.9154 &.9130 &.9097 &.9119  \\  
	
	evolgen		&.9133 &.9145 &.9157 &.9135 &.9098 &.9130  \\
	     		
    \cmidrule(r){2-7}
	Train labels avg	&.8835 	&.8806 	&.8991 	&.8739 	&.8791 	&.8847 	\\

    \bottomrule
\end{tabular}
}
\caption{\label{tab:chalearn_table} Mean accuracy for each individual modality, fusion approaches, two winners of the ChaLearn 2016 competition (DCC and evolgen), and average of training set labels. }
\end{table}

\subsection{Discussion}
Looking at the results obtained from various fusion methods in Table~\ref{tab:cnn_tri-modal_table}, we can see that for decision-level and last-layer backpropagation, Neuroticism and Extraversion are the easiest traits to predict, followed by Conscientiousness and Openness. Agreeableness is significantly harder. We also see that the last-layer fusion yields very similar performance as the decision-level approach. It is likely that the limited backpropagation method learns something similar to a linear combination of channels, just like the decision-level method. On the other hand, the full backpropagation method yields significantly higher results for all traits except Agreeableness. 

From Table~\ref{tab:tri-modal-weights} we can also see which modalities carry more information. The text modality is not adding much value to traits other than Agreeableness and Conscientiousness. Extraversion can, on the other hand, be quite easily recognized from tone of voice and appearance. Having said this, we must be careful in deciding which modalities are most suitable for individual traits, since certain traits (e.g. Extraversion) are more evident from a short slice and some (e.g. Openness) need longer temporal information~\cite{aran2013one}. \newcite{polzehl2010automatically} have proposed a method for personality recognition in speech modality on a different corpus. The method is not based on neural network architecture, but they provide a similar analysis that supports our conclusions.

Since the full backpropagation experiments yields much better results than the linear combination model, we can conclude that different modalities interact with each other in a non-trivial manner. Moreover, we can observe that simply adding features from different modalities (represented as concatenating a final representation without full backpropagation) does not yield optimal performance.

Our tri-modal approach is quite extensive and there are more modalities such as nationality, cultural background, age, gender, and personal interests that can be added. All Big Five traits have been found to have a correlation with age~\cite{donnellan2008age}. Extraversion and Openness have a negative correlation with age, Agreeableness have a positive correlation, and Conscientiousness scores peak for middle age subjects.

\section{Conclusion} \label{sec:conclusion}
We proposed a fusion method, based on deep neural networks, to predict personality traits from audio, language and appearance. We have seen that each of the three modalities contains a signal relevant for personality prediction, that using all three modalities combined greatly outperforms using individual modalities, and that the channels interact with each other in a non-trivial fashion. By combining the last network layers and fine-tuning the parameters we have obtained the best result, average among all traits, of 0.0938 Mean Square Error, which is 9.4\% better than the performance of the best individual modality (visual). Out of all modalities, language or speech pattern seems to be the least relevant. Video frames (appearance) are slightly more relevant than audio information (i.e. non-verbal parts of speech). 
\section{Acknowledgments}
This work was partially funded by grants \#16214415 and \#16248016 of the Hong Kong Research Grants Council, ITS/319/16FP of Innovation Technology Commission, and RDC 1718050-0 of EMOS.AI.

\bibliography{acl2018}
\bibliographystyle{acl_natbib}

\end{document}